\documentclass[10pt,twocolumn,letterpaper]{article}

\usepackage{cvpr}              

\usepackage{graphicx}
\usepackage{amsmath}
\usepackage{amssymb}
\usepackage{booktabs}
\usepackage{multirow}
\usepackage[accsupp]{axessibility}

\usepackage[pagebackref,breaklinks,colorlinks]{hyperref}

\usepackage[capitalize]{cleveref}
\crefname{section}{Sec.}{Secs.}
\Crefname{section}{Section}{Sections}
\Crefname{table}{Table}{Tables}
\crefname{table}{Tab.}{Tabs.}

\def\cvprPaperID{851} 
\def\confName{CVPR}
\def\confYear{2023}

\begin{document}

\title{Phase-Shifting Coder: Predicting Accurate Orientation\\ in Oriented Object Detection}
              
\author{Yi Yu$^{1,2}$, Feipeng Da$^{1,2,}$\thanks{Corresponding author is Feipeng Da. This work is supported by Special Project on Basic Research of Frontier Leading Technology of Jiangsu Province of China (BK20192004C).}\\
$^{1}$School of Automation, Southeast University, Nanjing, China\\
$^{2}$Key Laboratory of Measurement and Control of Complex
Systems of Engineering,\\ Ministry of Education, Southeast University, Nanjing, China\\
{\tt\small \{yuyi, dafp\}@seu.edu.cn}
}
\maketitle

\begin{abstract}
With the vigorous development of computer vision, oriented object detection has gradually been featured. In this paper, a novel differentiable angle coder named phase-shifting coder (PSC) is proposed to accurately predict the orientation of objects, along with a dual-frequency version (PSCD). By mapping the rotational periodicity of different cycles into the phase of different frequencies, we provide a unified framework for various periodic fuzzy problems caused by rotational symmetry in oriented object detection. Upon such a framework, common problems in oriented object detection such as boundary discontinuity and square-like problems are elegantly solved in a unified form. Visual analysis and experiments on three datasets prove the effectiveness and the potentiality of our approach. When facing scenarios requiring high-quality bounding boxes, the proposed methods are expected to give a competitive performance. The codes are publicly available at https://github.com/open-mmlab/mmrotate.
\end{abstract}

\section{Introduction}
\label{sec:intro}

As a fundamental task in computer vision, object detection has been extensively studied. Early researches are mainly focused on horizontal object detection\cite{Zhao2019Object}, on the ground that objects in natural scenes are usually oriented upward due to gravity. However, in other domains such as aerial images \cite{Xia2018DOTA,Liu2017HRSC,Yang2018Automatic,Yang2022Arbitrary,Fu2020Rotation}, scene text \cite{Nayef2017ICDAR,Ma2018Arbitrary,Liao2018Rotation,Liu2018FOTS,Zhou2017EAST}, and industrial inspection \cite{Yu2022OCDPCB,Liu2020Data,Wu2022PCBNet}, oriented bounding boxes are considered more preferable. Upon requirements in these scenarios, oriented object detection has gradually been featured.

At present, several solutions around oriented object detection have been developed, among which the most intuitive way is to modify horizontal object detectors by adding an output channel to predict the orientation angle. Such a solution faces two problems:

1) Boundary problem \cite{Yang2020Arbitrary}: Boundary discontinuity problem is often caused by angular periodicity. Assuming orientation $-\pi/2$ is equivalent to $\pi/2$, the network output is sometimes expected to be $-\pi/2$ and sometimes $\pi/2$ when facing the same input. Such a situation makes the network confused about in which way it should perform regression.

2) Square-like problem \cite{Yang2021Rethinking}: Square-like problem usually occurs when a square bounding box cannot be uniquely defined. Specifically, a square box should be equivalent to a ${90}^{\circ}$ rotated one, but the regression loss between them is high due to the inconsistency of angle parameters. Such ambiguity can also seriously confuse the network.

A more comprehensive introduction to these problems can be found in previous researches \cite{Yang2022Arbitrary, Yang2021Rethinking}. Also, several methods have been proposed to address these problems, which will be reviewed in \cref{sec:relatedwork}. 

Through rethinking the above problems, we find that they can inherently be unified as rotationally symmetric problems (boundary under ${180}^{\circ}$ and square-like under ${90}^{\circ}$ rotation), which is quite similar to the periodic fuzzy problem of the absolute phase acquisition \cite{Zuo2016Temporal} in optical measurement. Inspired by this, we come up with an idea to utilize phase-shifting coding, a technique widely used in optical measurement \cite{Zuo2018Phase}, for angle prediction in oriented object detection. The technique has the potential to solve both boundary discontinuity and square-like problems:

1) Phase-shifting encodes the measured distance (or parallax) into the periodic phase in optical measurement. The orientation angle can also be encoded into the periodic phase, and boundary discontinuity is thus inherently solved.

2) Phase-shifting also has the periodic fuzzy problem, which is similar to the square-like problem, and many solutions exist. For example, the dual-frequency phase-shifting technique solves the periodic fuzzy problem by mixing phases of different frequencies (also known as phase unwrapping \cite{Zuo2016Temporal}). 

\textbf{Motivations of this paper:}

Based on the above analysis, we believe that the phase-shifting technique can be modified and adapted to oriented object detection. What is the principle of the phase-shifting angle coder? How to integrate this module into a deep neural network? Will this technique result in better performance? These questions are what this paper is for.

\textbf{Contributions of this paper:}

1) We are the first to utilize the phase-shifting coder to cope with the angle regression problem in the deep learning area. An integral and stable solution is elaborated on in this paper. Most importantly, the codes are well-written, publicly available, and with reproducible results.

2) The performance of the proposed methods is evaluated through extensive experiments. The experimental results are of high quality---All the listed results are retested on identical environments to ensure fair comparisons (instead of copied from other papers).

\textbf{The rest of this paper is organized as follows:}

Section 2 reviews the related methods around oriented object detection. Section 3 describes the principles of the phase-shifting coder in detail. Section 4 conducts experiments on several datasets to evaluate the performance of the proposed methods. Section 5 concludes the paper.

\section{Related work}
\label{sec:relatedwork}

With datasets such as DOTA \cite{Xia2018DOTA}, HRSC \cite{Liu2017HRSC}, and ICDAR \cite{Nayef2017ICDAR}, extensive studies around oriented object detection have been carried out, and some representative ones are summarized in this section. 

\subsection{From horizontal to oriented}

Many oriented object detection methods are based on horizontal object detection, which has been reviewed in the literature \cite{Ren2017Faster, Lin2017Feature, Zhao2019Object}. Hence, we only briefly introduce four representative frameworks:

1) \textbf{\emph{Anchor-based.}} RetinaNet (2017) \cite{Lin2017Focal}: As a one-stage detector, RetinaNet uses Feature Pyramid Network (FPN) as the backbone, to which two subnetworks are attached, one for classification and the other for regression.

2) \textbf{\emph{Anchor-free.}} FCOS (2019) \cite{Tian2019FCOS}: By adding a center-ness regression, this work presents an anchor-free one-stage detector, which avoids the complicated computation and hyper-parameters related to anchor boxes.

3) \textbf{\emph{Point-based.}} RepPoint (2019) \cite{Yang2019Reppoints}: RepPoint is a representation of objects as a set of sample points, which learn to arrange themselves in a manner that bounds the spatial extent of an object and indicates significant local areas.

4) \textbf{\emph{High-efficiency.}} YOLO (v5, 2021) \cite{Jocher2021YOLOv5}: Famous for its high speed and accuracy, YOLO divides images into a grid system. Each cell in the grid is responsible for detecting objects within itself.

By adding an output channel to predict the orientation of each object, these horizontal detectors can comfortably be applied to oriented object detection, usually termed as \emph{Rotated RetinaNet}, \emph{Rotated FCOS}, \emph{Rotated RepPoint}, and \emph{Rotated YOLO} \cite{Zhou2022MMRotate}. 

\subsection{Rotation-invariant detectors}

Based on the above frameworks, some studies use additional modules to cope with rotation and improve performance. Some representative ones are as follows:

1) RoI Transformer (2019) \cite{Ding2018Learning}: The core idea is to apply spatial transformations on Regions of Interest (RoIs) and learn the transformation parameters under the supervision of oriented bounding box (OBB) annotations.

2) ReDet (2021) \cite{Han2021Redet}: Rotation-equivariant Detector (ReDet) is proposed in this work to explicitly encode rotation equivariance and rotation invariance. A rotation-equivariant network is used to accurately predict the orientation, and upon that, rotation-invariant features can be extracted through rotation-invariant RoI align.

3) S2ANet (2021) \cite{Han2021Align}: Single-shot alignment network (S2ANet) consists of two modules: a feature alignment module (FAM) and an oriented detection module (ODM). The FAM can generate high-quality anchors, and the ODM adopts active rotating filters to encode the orientation information and produce orientation-invariant features.

4) R3Det (2021) \cite{Yang2021R3Det}: An end-to-end refined one-stage rotation detector is proposed for fast and accurate object detection by using a progressive regression approach from coarse to fine granularity, followed by a feature refinement module to further improve detection performance.

5) Oriented R-CNN (2021) \cite{Xie2021Oriented}: This work proposes a general two-stage oriented detector with promising accuracy and efficiency. In the first stage, an oriented Region Proposal Network (oriented RPN) directly generates high-quality oriented proposals in a nearly cost-free manner. The second stage is oriented R-CNN head for refining oriented Regions of Interest (oriented RoIs) and recognizing them.

Most of the above methods enhance the network architecture and improve the rotation invariance of the output features. With the progress in related fields, some researchers found that to further improve the performance, two other problems need to be concerned: \textit{boundary discontinuity} and \textit{square-like} \cite{Yang2020Arbitrary, Yang2021Rethinking}.

\begin{figure*}[t]
\setlength{\abovecaptionskip}{0mm}
\centering
\includegraphics[width=0.85\linewidth]{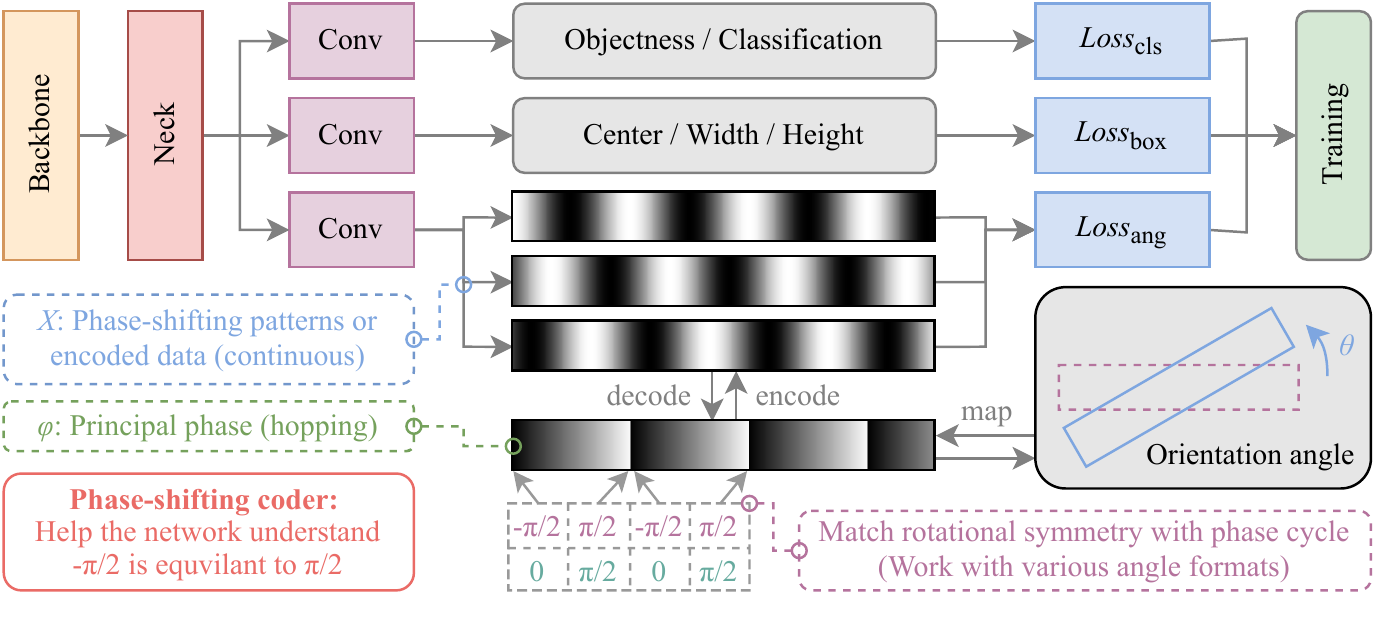}
\caption{Overall flowchart of the phase-shifting coder integrated into a deep neural network.}
\label{fig:1}
\end{figure*}

\subsection{Boundary and square-like problems}

Since these two problems were pinpointed, many techniques have been proposed to solve part of them, which can be divided into three categories:

1) \textbf{\emph{Smooth loss function.}} SCRDet (2019) \cite{Yang2019SCRDet}: IoU-smooth L1 loss is proposed to smooth the boundary loss jump; RSDet (2021) \cite{Qian2021RSDet}: Similarly, a modified version using modulated loss is proposed in this work.

2) \textbf{\emph{Angle coder.}} CSL (2020) \cite{Yang2020Arbitrary}: Circular smooth label (CSL) technique converts angle regression to classification to handle the periodicity of the orientation angle and increase the error tolerance to adjacent angles; Upon CSL,
DCL \cite{Yang2021Dense} further solves the square-like problem.

3) \textbf{\emph{Gaussian distribution.}} GWD (2021) \cite{Yang2021Rethinking}: Regression loss based on Gaussian Wasserstein distance (GWD) is proposed, where the rotated bounding box is converted to a 2D Gaussian distribution to calculate the regression loss; KLD (2021) \cite{Yang2021Learning}: The Kullback-Leibler Divergence (KLD) between the Gaussian distributions is calculated as the regression loss; KFIoU (2022) \cite{Yang2022KFIoU}: Kalman filter is adopted to mimic the mechanism of Skew Intersection over Union (SkewIoU) by its definition, which requires less hyper-parameter tuning than GWD and KLD.

These methods consider the problems from different perspectives and each has its pros and cons. For example, SCRDet and RSDet are designed to alleviate the impact of the problems, instead of theoretically solving them; CSL is simple and stable, but not able to solve the square-like problem, and its performance could be greatly affected by hyper-parameters; GWD and KLD solve both problems elegantly, but their prediction is relatively inaccurate, resulting in high $\text{mAP}_{50}$ but low $\text{mAP}_{75}$ performance.

Based on the above analysis, much progress has been made in the field of oriented object detection, but related problems have not yet been completely solved.

\section{Method}
\label{sec:method}

In this section, we will first introduce the encoding and decoding procedure of the phase-shifting coder. Afterward, an enhanced version dual-frequency phase-shifting coder will be introduced. Readers are referred to literature \cite{Zuo2018Phase} for the basic principles of the phase-shifting technique.

\subsection{Phase-shifting coder (PSC)}

\begin{figure*}[t]
\setlength{\abovecaptionskip}{1.5mm}
\centering
\includegraphics[width=0.87\linewidth]{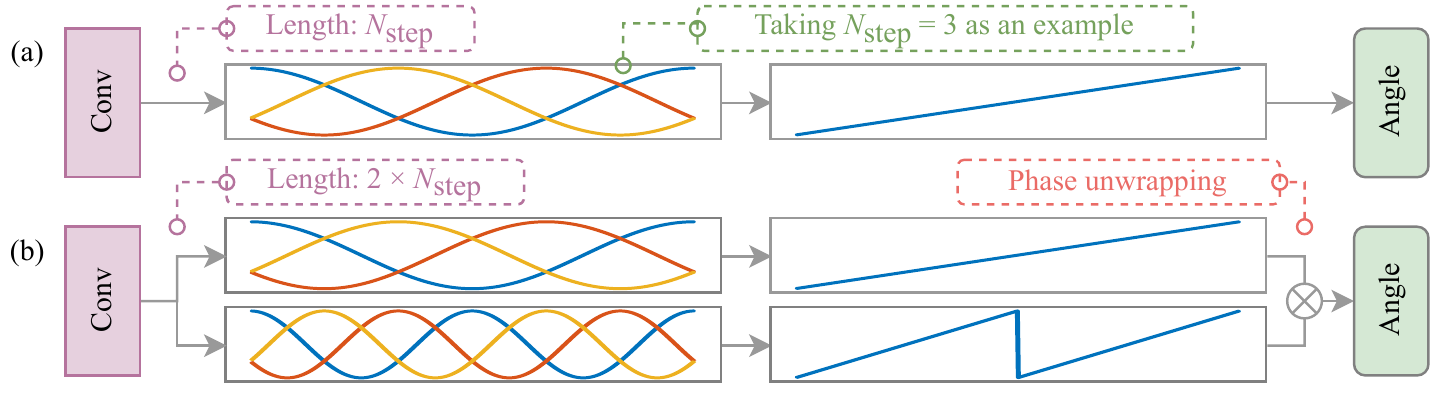}
\caption{Inference flow of (a) single-frequency PSC and (b) dual-frequency PSC.}
\label{fig:2}
\end{figure*}

The overall flowchart of the phase-shifting coder integrated into a deep neural network is illustrated in \cref{fig:1}. Taking the ``long edge 90" angle definition as an example for illustration, symbols can be defined as follows:

\begin{itemize}
\setlength{\itemsep}{5pt}
\setlength{\parsep}{0pt}
\setlength{\parskip}{0pt}
\item $\theta$: Orientation angle, in range $\left [ -\pi/2  , \pi/2 \right )$
\item $\varphi$: Principal phase, in range $\left [ -\pi  , \pi \right )$
\item $N_{\text {step}}$: The number of phase-shifting steps
\item $X$: Encoded data, $X=\left\{x_{n} \mid n=1,2, \cdots, N_{\text {step}}\right\}$
\end{itemize}

\textbf{Mapping:}
The cycle of $\sin$ or $\cos$ is $2\pi$, whereas a rectangle box is identical to itself when rotated by $\pi$, thus a mapping is required to match them, as follows:

\begin{equation}
\varphi = 2 \theta
\end{equation}

\textbf{Encoding:}
The formula of encoding $\varphi$ into $X$ can be described as:

\begin{equation}
x_{n}= \cos \left(\varphi+\frac{2 n \pi }{ N_\text{step}}\right ) 
\label{eq:fenc}
\end{equation}
where $n=1,2, \cdots, N_{\text {step}}$. 

To simplify the subsequent description, \cref{eq:fenc} is also denoted as $X=f_\text{enc}(\varphi)$.

\textbf{Decoding:}
The formula of decoding $\varphi$ from $X$ can be described as:

\begin{equation}
\varphi=-\arctan \frac{\sum_{n=1}^{N_\text{step}} x_{n} \sin \left(\frac{2 n \pi}{N_\text{step}}\right)}{\sum_{n=1}^{N_\text{step}} x_{n} \cos \left(\frac{2 n \pi}{N_\text{step}}\right)}
\label{eq:fdec}
\end{equation}

The $\arctan $ in the formula should be implemented by the arctan2 function so that its output is in the range $\left ( -\pi, \pi \right ]$. \Cref{eq:fdec} is also denoted as $\varphi=f_\text{dec}(X)$.

\subsection{Dual-frequency phase-shifting coder (PSCD)}

Through rethinking the boundary problem and the square-like problem, we believe these two problems can be inherently unified. If a bounding box is equivalent to itself under ${180}^{\circ}$ rotation, the boundary problem occurs, but if they are equivalent under ${90}^{\circ}$ rotation, the square-like problem occurs. Both cases are periodic fuzzy problems but of different cycles.

Therefore, to solve both boundary discontinuity and square-like problems, an additional phase is required to establish the dual-frequency phase-shifting coder. The difference between the basic phase-shifting coder and the dual-frequency one is illustrated in \cref{fig:2}. 

Additional symbols used in dual-frequency phase-shifting coder can be defined as follows:

\begin{itemize}
\setlength{\itemsep}{5pt}
\setlength{\parsep}{0pt}
\setlength{\parskip}{0pt}
\item $\varphi_1$: Phase of the first frequency, in range $\left [ -\pi  , \pi \right )$
\item $\varphi_2$: Phase of the second frequency, in range $\left [ -2\pi  , 2\pi \right )$
\item $\varphi$: Final principal phase, in range $\left [ -\pi  , \pi \right )$ 
\item $X_1$: Data encoded from the phase of the first frequency, $X_1=\left\{x_{n} \mid n=1,2, \cdots, N_{\text {step}}\right\}$
\item $X_2$: Data encoded from the phase of the second frequency, $X_2=\left\{x_{n} \mid n=1,2, \cdots, N_{\text {step}}\right\}$
\item $X$: Final encoded data, with coding length $2\times N_{\text {step}}$, $X=\left\{X_1, X_2\right\}$
\end{itemize}

\textbf{Mapping:}
In dual-frequency PSC, two principal phases are mapped from angle $\theta$ during the training process:

\begin{equation}
\left\{\begin{matrix}
\varphi_1 = 2 \theta \\
\varphi_2 = 4 \theta
\end{matrix}\right.
\end{equation}

The output orientation angle is mapped from the final principal phase during the inference process:

\begin{equation}
\theta = \frac{1}{2} \varphi
\end{equation}

\textbf{Encoding:}
Similar to \cref{eq:fenc}, the formula of encoding $\varphi_1$ and $\varphi_2$ into $X_1$ and $X_2$ can be described as:

\begin{equation}
\left\{\begin{matrix}
X_1=f_\text{enc}(\varphi_1) \\
X_2=f_\text{enc}(\varphi_2)
\end{matrix}\right.
\end{equation}

\textbf{Decoding:}
Similar to \cref{eq:fdec}, the formula of decoding $\varphi_1$ and $\varphi_2$ from $X_1$ and $X_2$ can be described as:

\begin{equation}
\left\{\begin{matrix}
\varphi_1=f_\text{dec}(X_1) \\
\varphi_2=f_\text{dec}(X_2)
\end{matrix}\right.
\end{equation}

\textbf{Unwrapping:}
The network outputs two principal phases during inference: $\varphi_1$ as the absolute phase and $\varphi_2$ as the wrapped phase. We need to mix them to obtain the final phase (also known as phase unwrapping). To this end, we first calculate the inner product between $\varphi_1$ and $\varphi_2$ by:

\begin{equation}
\delta  = \cos \varphi_1 \cos \frac{\varphi_2}{2} + \sin \varphi_1 \sin \frac{\varphi_2}{2}
\end{equation}

Afterward, $\varphi_2$ is unwrapped according to $\delta$, so that the two phases can be automatically mixed to obtain the final phase $\varphi $, as follows:

\begin{equation}
\varphi = \begin{cases}
\pi +\frac{\varphi_2}{2}, & \text{if}\ \ \delta<0\\
\frac{\varphi_2}{2}, & \text{else}
\end{cases}
\end{equation}

It should be noted that the above formula is a simplified version for higher clarity. In fact, after added by $\pi$, $\varphi$ could be out of the range $\left [ -\pi  , \pi \right )$. In such case, $\varphi$ needs to be subtracted by $2 \pi$, otherwise the output angle could be outside the definition range.

\subsection{General form of mapping}

\begin{figure}[ht]
\setlength{\abovecaptionskip}{1mm}
\centering
\includegraphics[width=0.82\linewidth]{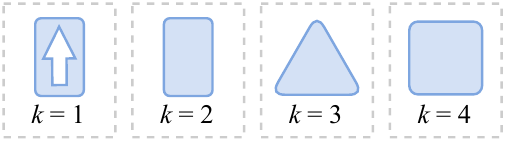}
\caption{Objects with different rotational periodicity.}
\label{fig:5}
\end{figure}

Similarly, the ambiguity caused by ``triangle-like" or ``pentagon-like" objects can also be solved, as well as the ${360}^{\circ}$ orientation problem (for example, predicting the heading direction of the airplane), by extending the mapping between phase and angle to a more general form:

\begin{equation}
\varphi = k \theta
\end{equation}
where $k = 2\pi/s $, assuming objects to be detected are symmetric under the rotation of $s$ radians, as shown in \cref{fig:5}.

PSC is applicable in all of these cases, while the multi-frequency strategy can be used when different types exist simultaneously. In other words, by mapping the rotational periodicity of different cycles into the phase of different frequencies, we provide a unified framework for various rotational symmetry problems in oriented object detection.

\subsection{Loss function}

The oriented bounding box is represented by five parameters $(x, y, w, h, \theta)$, denoting the box’s center coordinates,
width, height, and angle, respectively. As an angle-coder-based method, PSC only involves the regression of $\theta$.

According to \cref{eq:fenc}, the encoded data (phase-shifting patterns) are in the range $\left [ -1, 1 \right ]$. To make the training more stable, we also transform the output features with:

\begin{equation}
X_\text{Pred} = 2\times\text{sigmoid}(X_\text{Feat}) - 1
\end{equation}
where $X_\text{Feat}$ is the output features of the convolution layer, and $X_\text{Pred}$ is the predicted encoded data in range $\left [ -1  , 1 \right ]$.

Afterward, the loss of the angle branch can be calculated with L1 loss:

\begin{equation}
L_\text{ang}=\left | X_\text{GT} - X_\text{Pred} \right |
\end{equation}
where $X_\text{GT}$ is the ground truth phase-shifting patterns encoded from the orientation angle of annotated boxes.

Finally, the overall loss can be expressed as:

\begin{equation}
L=w_1L_\text{cls} + w_2L_\text{box} + w_3L_\text{ang}
\label{eq:lossall}
\end{equation}
where $w_1L_\text{cls}$ and $w_2L_\text{box}$ are the weighted loss of classification and box regression branches defined by the backbone detector, and $w_3$ is set to $0.2w_1$ by default. 

It should be noted that \cref{eq:lossall} describes only the general situation, and there are also some special cases, such as FCOS with center-ness loss.

\section{Experiments}
\label{sec:experiments}

With the help of PyTorch \cite{Paszke2019PyTorch}, ultralytics/yolov5 \cite{Jocher2021YOLOv5}, and MMRotate \cite{Zhou2022MMRotate} tool kits, experiments are carried out to evaluate the performance of the proposed methods. To compare with existing literature, we choose mean average precision (mAP) as the major metric. The computing infrastructure is as follows: CPU: Intel i9-12900K, GPU: Nvidia RTX3080, OS: Windows 10, PyTorch: 1.10.1, ultralytics/yolov5: 6.0, MMRotate: 0.3.2.

\subsection{Datasets and benchmarks}

\textbf{DOTA \cite{Xia2018DOTA}:} DOTA is comprised of 2,806 large aerial images---1,411 for training, 937 for validation, and 458 for testing. The dataset is annotated using 15 categories with 188,282 instances in total. The categories are defined as: Plane (PL), Baseball Diamond (BD), Bridge (BR), Ground Field Track (GTF), Small Vehicle (SV), Large Vehicle (LV), Ship (SH), Tennis Court (TC), Basketball Court (BC), Storage Tank (ST), Soccer-Ball Field (SBF), Roundabout (RA), Harbor (HA), Swimming Pool (SP), and Helicopter (HC). We follow the standard preprocessing procedure in MMRotate---The high-resolution images are split into 1024 $\times$ 1024 patches with an overlap of 200 pixels for training, and during inference, the detection results of all patches are merged to evaluate the performance.

\textbf{HRSC \cite{Liu2017HRSC}:} As a ship detection dataset, HRSC contains ship instances both on the sea and inshore, with arbitrary orientation. The training, validation, and testing set include 436, 181, and 444 images respectively. We use the preprocessing provided by MMRotate, where the images are scaled to 800 $\times$ 800 for training and testing.

\textbf{OCDPCB \cite{Yu2022OCDPCB}:} OCDPCB is a dataset for oriented component detection in printed circuit boards aimed at automated optical inspection. The dataset consists of 636 images, of which 445 images are used for training and 191 for testing. The resolution of the images is 1280 $\times$ 1280.

\subsection{Ablation study}

\begin{figure*}[t]
\setlength{\abovecaptionskip}{0.5mm}
\centering
\includegraphics[width=0.92\linewidth]{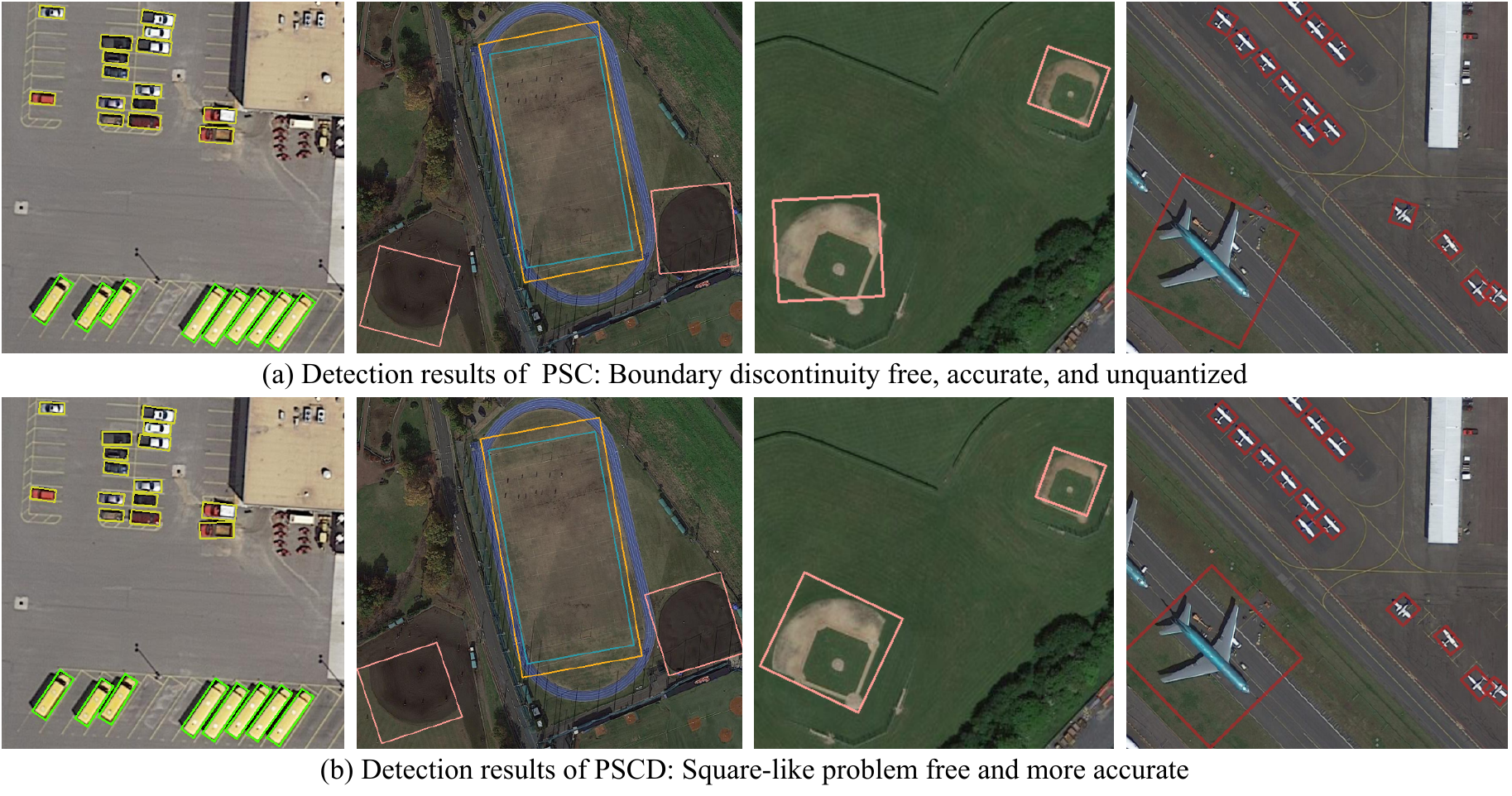}
\caption{Visual comparison between single-frequency PSC and dual-frequency PSC.}
\label{fig:4}
\end{figure*}

\textbf{Ablation study of hyper-parameters:}

In most existing methods such as CSL, GWD, and KLD, hyper-parameters could highly affect the performance. Worse still, the best parameters vary in different scenarios and datasets, which require laborious tuning. 

PSC is quite different---The only hyper-parameter $N_{\text {step}}$ is an integer greater than or equal to 3, so we evaluate several $N_{\text {step}}$ values most commonly used in the phase-shifting technique, and the results, which are obtained on RetinaNet with dual-frequency PSC, are shown in \cref{tab:nstep}.

\begin{table}[ht]
\centering
\small
\renewcommand{\arraystretch}{1.2}
\setlength{\tabcolsep}{2mm}
\setlength{\abovecaptionskip}{2mm}
\begin{tabular}{c|c|c|c}
\hline
\multicolumn{1}{c|}{Metrics} & $N_\text{step} = 3$ & $N_\text{step} = 4$ & $N_\text{step} = 5$ \\ \hline\hline
DOTA ($\text{mAP}_{50}$) & {\color{red}\bf71.09}             & 70.96             & 71.01             \\ \hline
DOTA ($\text{mAP}_{75}$) & 41.17             & {\color{red}\bf41.82}             & 41.53             \\ \hline
DOTA ($\text{mAP}_{50:95}$) & 41.25             & {\color{red}\bf41.51}             & 41.13             \\ \hline\hline
HRSC ($\text{mAP}_{50}$) & {\color{red}\bf85.53}             & 85.49             & 85.46             \\ \hline
HRSC ($\text{mAP}_{75}$) & 59.57             & {\color{red}\bf59.64}             & 59.54             \\ \hline
HRSC ($\text{mAP}_{50:95}$) & 53.20             & {\color{red}\bf53.22}             & 53.12             \\ \hline
\end{tabular}
\caption{Performance under different $N_\text{step}$ value.}
\label{tab:nstep}
\end{table}

According to the experimental results, this parameter has quite a limited impact on the results. Although $N_{\text {step}} = 4$ shows better $\text{mAP}_{75}$ performance, it also impairs $\text{mAP}_{50}$ performance. On the whole, larger $N_{\text {step}}$ will not bring significant benefits. Taking the computational complexity into consideration, we would recommend $N_{\text {step}} = 3$. 

Once $N_{\text {step}} = 3$ is determined, PSC has no adjustable hyper-parameters.

\textbf{Ablation study of dual-frequency:}

We provide an intuitive comparison between single-frequency PSC and dual-frequency PSC to verify the effectiveness of the dual-frequency module and help researchers decide which one to choose.

The visual comparison displayed in \cref{fig:4} quite conforms with our theory, proving that the dual-frequency strategy can work as expected and solve both boundary discontinuity and square-like problems in a unified way. And thus, the dual-frequency strategy is highly recommended in scenarios containing square-like objects.


\begin{table*}
\centering
\footnotesize
\renewcommand{\arraystretch}{1.43}
\setlength{\tabcolsep}{1.0mm}
\setlength{\abovecaptionskip}{2mm}
\begin{tabular}{l|l|c|c|c|c|c|c|c|c|c|c|c|c|c|c|c|c|c|c|c}
\hline
\multicolumn{2}{c|}{Method}            & R & PL    & BD    & BR    & GTF   & SV    & LV    & SH    & TC    & BC    & ST    & SBF   & RA    & HA    & SP    & HC    & \scalebox{0.9}[1.0]{$\text{mAP}_{50}$} & \scalebox{0.9}[1.0]{$\text{mAP}_{75}$} & \scalebox{0.65}[1.0]{$\text{mAP}_{50:95}$} \\ \hline
\hline
\multirow{5}{*}{\,\rotatebox{90}{FCOS (R-50)}\,} & Rotated &    & {\color{red}\bf89.18} & 71.99 & 47.97 & 61.61 & 79.30 & 73.52 & 85.78 & {\color{red}\bf90.90} & 81.09 & 84.30 & {\color{red}\bf59.57} & {\color{red}\bf62.69} & 62.08 & 69.94 & 49.31 & 71.28 & 37.08 & 39.42    \\ \cline{2-21}
& +KLD          &    & {\color{blue}89.17} & {\color{red}\bf75.42} & {\color{red}\bf49.41} & 56.48 & 79.66 & 76.78 & 86.91 & {\color{blue}90.89} & {\color{red}\bf83.52} & {\color{blue}84.41} & 58.76 & 62.21 & 63.44 & 67.90 & 50.07 & {\color{blue}71.67} & 37.53 & 39.67    \\ \cline{2-21}
& +CSL          &    & 88.24 & {\color{blue}74.87} & 41.27 & 61.03 & 79.52 & 78.35 & 87.19 & 90.88 & 81.50 & {\color{red}\bf84.53} & 54.70 & {\color{blue}62.65} & 62.84 & 68.45 & 46.50 & 70.83 & 38.71 & 39.75    \\ \cline{2-21}
& +PSC          &    & 88.24 & 74.42 & 48.63 & {\color{blue}63.44} & {\color{blue}79.98} & {\color{red}\bf80.76} & {\color{red}\bf87.59} & 90.88 & {\color{blue}82.02} & 71.58 & {\color{blue}59.12} & 60.78 & {\color{red}\bf65.78} & {\color{red}\bf71.21} & {\color{blue}53.06} & {\color{red}\bf71.83} & {\color{blue}39.21} & {\color{red}\bf40.42}    \\ \cline{2-21}
& +PSCD         &    & 88.04 & 73.95 & {\color{blue}48.83} & {\color{red}\bf63.44} & {\color{red}\bf80.01} & {\color{blue}80.75} & {\color{blue}87.58} & 90.88 & 81.69 & 67.23 & 58.70 & 60.26 & {\color{blue}65.67} & {\color{blue}71.11} & {\color{red}\bf53.06} & 71.41 & {\color{red}\bf39.35} & {\color{blue}40.36}    \\ \hline
\hline
\multirow{4}{*}{\,\rotatebox{90}{FCOS (R-50)}\,} & +KLD          & \checkmark & 89.05 & {\color{red}\bf 73.73} & {\color{red}\bf 49.17} & 57.86 & {\color{red}\bf 79.54} & {\color{red}\bf 77.91} & 87.41 & {\color{blue} 90.87} & {\color{red}\bf 83.23} & {\color{blue} 82.42} & 58.97 & 61.83 & 62.91 & 72.69 & 59.44 & 72.47 & 37.84 & 39.95    \\ \cline{2-21}
& +CSL          & \checkmark & 87.84 & 68.88 & 42.15 & 57.41 & {\color{blue} 77.24} & 74.78 & 87.66 & {\color{red}\bf 90.89} & 79.43 & {\color{red}\bf 84.34} & 54.35 & 63.46 & 61.39 & 69.62 & 61.16 & 70.71 & 37.13 & 38.47    \\ \cline{2-21}
& +PSC          & \checkmark & {\color{red}\bf 89.07} & 73.60 & 48.91 & {\color{red}\bf 62.63} & 75.24 & {\color{blue} 77.71} & {\color{red}\bf 88.02} & 90.85 & {\color{blue} 82.86} & 69.85 & {\color{red}\bf 61.48} & {\color{red}\bf 65.25} & {\color{blue} 65.68} & {\color{red}\bf 73.03} & {\color{red}\bf 67.88} & {\color{blue} 72.80} & {\color{blue} 38.83} & {\color{blue} 41.10}    \\ \cline{2-21}
& +PSCD         & \checkmark & {\color{blue} 89.06} & {\color{blue} 73.61} & {\color{blue} 49.03} & {\color{blue} 62.34} & 75.18 & 77.69 & {\color{blue} 88.00} & 90.85 & 82.63 & 72.97 & {\color{red}\bf 61.48} & {\color{blue} 64.20} & {\color{red}\bf 65.77} & {\color{blue} 72.88} & {\color{red}\bf 67.88} & {\color{red}\bf 72.90} & {\color{red}\bf 39.80} & {\color{red}\bf 41.51}    \\ \hline
\hline
\multirow{5}{*}{\,\rotatebox{90}{RetinaNet (R-50)}\,} & Rotated &    & 89.41 & 76.81 & {\color{red}\bf 40.88} & 67.54 & 77.51 & 62.63 & 77.55 & {\color{blue} 90.89} & 82.31 & 81.98 & 58.16 & 61.56 & 56.46 & 63.71 & 38.96 & 68.42 & {\color{red}\bf 42.03} & {\color{blue} 40.13}     \\ \cline{2-21}
& +KLD     &    & {\color{red}\bf 89.50} & 79.91 & 39.92 & {\color{blue} 70.40} & {\color{blue} 78.04} & {\color{blue} 64.24} & {\color{red}\bf 82.79} & {\color{red}\bf 90.90} & 81.80 & {\color{red}\bf 83.02} & 57.63 & 63.52 & 56.63 & {\color{blue} 65.13} & {\color{blue} 50.04} & {\color{blue} 70.23} & 37.88 & 39.31     \\ \cline{2-21}
& +CSL     &    & 89.33 & 79.67 & {\color{blue} 40.83} & 69.95 & 77.71 & 62.08 & 77.46 & 90.87 & {\color{blue} 82.87} & {\color{blue} 82.03} & 60.07 & {\color{blue} 65.27} & 53.58 & 64.03 & 46.62 & 69.49 & 40.42 & 39.69     \\ \cline{2-21}
& +PSC     &    & {\color{blue} 89.41} & {\color{blue} 80.66} & 39.06 & 69.08 & 77.61 & 61.63 & 77.21 & 90.86 & 82.52 & 81.76 & {\color{red}\bf 60.98} & {\color{red}\bf 66.20} & {\color{blue} 57.51} & 64.75 & 48.28 & 69.83 & 40.37 & 40.03     \\ \cline{2-21}
& +PSCD    &    & 89.32 & {\color{red}\bf 82.29} & 37.92 & {\color{red}\bf 71.52} & {\color{red}\bf 78.40} & {\color{red}\bf 66.33} & {\color{blue} 78.01} & 90.89 & {\color{red}\bf 84.21} & 80.63 & {\color{blue} 60.22} & 64.73 & {\color{red}\bf 59.69} & {\color{red}\bf 68.37} & {\color{red}\bf 53.85} & {\color{red}\bf 71.09} & {\color{blue} 41.17} & {\color{red}\bf 41.25}    \\ \hline
\hline
\multirow{3}{*}{\,\rotatebox{90}{\scalebox{0.9}[1.0]{YOLOv5s}}\,} & +CSL       & \checkmark & 89.26 & 84.53 & 51.36 & {\color{blue} 60.69} & 80.70 & {\color{blue} 84.74} & {\color{blue} 88.41} & 90.68 & {\color{red}\bf 85.93} & 87.59 & 59.62 & 65.18 & 74.53 & 81.71 & {\color{red}\bf 66.91} & 76.79 & 46.85 & 45.63 \\ \cline{2-21}
& +PSC       & \checkmark & {\color{blue} 89.65} & {\color{blue} 86.37} & {\color{blue} 51.76} & {\color{red}\bf 63.42} & {\color{blue} 81.21} & 84.63 & 88.29 & {\color{blue} 90.80} & {\color{blue} 85.39} & {\color{blue} 87.93} & {\color{red}\bf 61.00} & {\color{blue} 66.41} & {\color{red}\bf 75.01} & {\color{blue} 81.77} & 66.20 & {\color{blue} 77.32} & {\color{blue} 47.56} & {\color{blue} 46.48} \\ \cline{2-21}
& +PSCD      & \checkmark & {\color{red}\bf 89.70} & {\color{red}\bf 86.68} & {\color{red}\bf 52.47} & 59.74 & {\color{red}\bf 81.57} & {\color{red}\bf 85.14} & {\color{red}\bf 88.44} & {\color{red}\bf 90.83} & 83.99 & {\color{red}\bf 88.40} & {\color{blue} 60.43} & {\color{red}\bf 69.04} & {\color{blue} 74.89} & {\color{red}\bf 83.30} & {\color{blue} 66.27} & {\color{red}\bf 77.40} & {\color{red}\bf 51.50} & {\color{red}\bf 48.27} \\ \hline
\hline
\multirow{3}{*}{\,\rotatebox{90}{\scalebox{0.9}[1.0]{YOLOv5m}}\,} & +CSL       & \checkmark & 89.60 & {\color{red}\bf 86.39} & 54.62 & {\color{red}\bf 62.04} & 80.36 & 85.20 & {\color{red}\bf 88.40} & {\color{red}\bf 90.80} & 80.54 & 88.47 & {\color{red}\bf 64.16} & 61.21 & 76.70 & {\color{red}\bf 83.13} & {\color{red}\bf 67.91} & 77.30 & 49.88 & 47.97 \\ \cline{2-21}
& +PSC       & \checkmark & {\color{blue} 89.85} & 85.93 & {\color{red}\bf 54.94} & 61.56 & {\color{blue} 81.89} & {\color{blue} 85.47} & 88.37 & {\color{blue} 90.73} & {\color{red}\bf 86.90} & {\color{blue} 88.79} & 63.90 & {\color{blue} 68.92} & {\color{blue} 76.82} & {\color{blue} 82.83} & {\color{blue} 63.25} & {\color{blue} 78.01} & {\color{blue} 50.50} & {\color{blue} 48.60} \\ \cline{2-21}
& +PSCD      & \checkmark & {\color{red}\bf 89.86} & {\color{blue} 86.02} & {\color{blue} 54.94} & {\color{blue} 62.02} & {\color{red}\bf 81.90} & {\color{red}\bf 85.48} & {\color{blue} 88.39} & 90.73 & {\color{blue} 86.90} & {\color{red}\bf 88.82} & {\color{blue} 63.94} & {\color{red}\bf 69.19} & {\color{red}\bf 76.84} & 82.75 & 63.24 & {\color{red}\bf 78.07} & {\color{red}\bf 54.10} & {\color{red}\bf 50.35} \\ \hline
\end{tabular}
\caption{$\text{AP}_{50}$ of each category and $\text{mAP}$ on DOTA. Column ``R" means using random rotation and random resize as augmentation.}
\label{tab:dota}
\end{table*}

\subsection{Experimental settings}

\textbf{Group settings:}

PSC can work with various detectors and backbones. Thus, we select three state-of-the-art backbones for experiments: FCOS (anchor-free), RetinaNet (anchor-based), and YOLO (high-efficiency). In the experiments, FCOS and RetinaNet are set to be based on ResNet-50 \cite{He2016Deep} (denoted as R-50), and YOLO includes two configurations: YOLOv5s and YOLOv5m. The parameter number of the four models is about FCOS (R-50): 32M, RetinaNet (R-50): 36M, YOLOv5s: 7M, and YOLOv5m: 21M.

We set up five experimental groups for DOTA dataset, and two for HRSC or OCDPCB datasets (as shown in \cref{tab:dota,tab:hrsc,tab:ocdpcb}). Both the backbone and the data augmentation are identical within each group to make the comparisons fair. Also, SWA \cite{Zhang2020SWA} and multi-scale testing are not adopted. Networks based on FCOS and RetinaNet are trained by 12 epochs on DOTA and 72 epochs on HRSC and OCDPCB, whereas the YOLO-based groups are trained by 120 epochs. The learning rate is initially set to 1e-3 and finally reduced to 1e-5. Readers are referred to the configuration files in our codes for all the particulars.

\textbf{Baseline methods:}

As an angle-coder-based method for boundary discontinuity and square-like problems, we take the following two most relevant methods as the baseline for comparison. 

(1) CSL \cite{Yang2020Arbitrary}: The most widely used angle coder converting the angle regression to classification to solve boundary discontinuity problem.

(2) KLD \cite{Yang2021Learning}: The state-of-the-art method solving both boundary discontinuity and square-like problems based on Gaussian distribution.

\subsection{Results and analysis}

The quantitative results on DOTA, HRSC, and OCDPCB datasets are demonstrated in \cref{tab:dota,tab:hrsc,tab:ocdpcb}. The top two results within each group are labeled in bold red and blue.

\textbf{Comparisons on DOTA dataset:}

1) \textbf{\emph{With KLD.}} PSC and PSCD show advantages over existing methods in many groups of experiments. Specifically, PSC is on a par with KLD in $\text{mAP}_{50}$, but significantly better than KLD in $\text{mAP}_{75}$. Such improvement in $\text{mAP}_{75}$ is even more significant for PSCD, which outperforms KLD by 2.36\,pp on average. Based on the above analysis, it can be concluded that PSC and PSCD show similar recall as KLD, but the bounding boxes of PSC and PSCD have higher Intersection over Union (IoU) and better quality.

2) \textbf{\emph{With CSL.}} In comparisons with CSL, both PSC and PSCD are superior in most metrics. On average, PSCD outperforms CSL by 1.15\,pp in $\text{mAP}_{50}$, and 2.59\,pp in $\text{mAP}_{75}$. It can be seen from the results that the gap between PSC and CSL is even more significant when data augmentation is used. We believe this phenomenon can be theoretically explained---By observing the training log, we find that the angle loss of PSC is much lower than that of CSL. With lower angle loss, PSC leaves more margin for the network to fit those augmented data, allowing the network to pay more attention to the classification and the bounding box branches, and finally resulting in higher performance. Furthermore, PSC is fully differentiable with unquantized outputs, which can be useful in detecting tiny deviations. This feature usually cannot be reflected by the mAP value, but could be important for some applications.

3) \textbf{\emph{Between PSC and PSCD.}} The $\text{mAP}_{75}$ of PSCD is considerably higher than that of PSC on DOTA dataset, with the improvement reaching 3.94\,pp on YOLOv5s and 3.60\,pp on YOLOv5m. In experiments using FCOS and RetinaNet backbones, PSCD also increases the $\text{mAP}_{75}$ by an average of 0.64\,pp compared with PSC.

\textbf{Comparisons on HRSC dataset:}

1) \textbf{\emph{With CSL and KLD.}} From the experimental results, the $\text{mAP}_{50}$ performance of different methods on HRSC dataset do not differ much, among which FCOS+PSC is the best, reaching 90.06 $\text{mAP}_{50}$. Whereas for $\text{mAP}_{75}$, there are noticeable differences among the methods---Although CSL achieves a decent $\text{mAP}_{50}$, its $\text{mAP}_{75}$ performance is significantly lower than other methods, indicating that the quality of the bounding boxes detected by CSL is relatively poor, with IoU generally below 75\%. When compared with the state-of-the-art method KLD, PSC also shows a significant advantage in $\text{mAP}_{75}$, with an improvement of 1.1\,pp and 2.54\,pp under FCOS and RetinaNet, respectively.

2) \textbf{\emph{Between PSC and PSCD.}} The dual-frequency strategy plays a negative role in some of the experiments. Such results of dual-frequency PSC, a technique mainly aimed at the square-like problem, are quite in line with our expectation---Objects in ship detection datasets are unlikely to be ``square-like". A similar phenomenon also occurs in OCDPCB dataset. These results provide guidance for researchers: PSC could be a better alternative than PSCD in scenarios containing rather few square-like objects.

\textbf{Comparisons on OCDPCB dataset:}

Most oriented object detectors have been well-tuned on DOTA and HRSC datasets, but when adapted to a completely new dataset, they might give a performance far below expectation. Therefore, we choose a less common dataset and train the networks with the same hyper-parameters used in training HRSC.

Surprisingly, without tuning hyper-parameters (such as the smooth radius in CSL) specifically for OCDPCB dataset, CSL and KLD produce rather limited effects, with $\text{mAP}_{50}$ slightly higher but $\text{mAP}_{75}$ much lower under RetinaNet, and negative results under FCOS.

PSC and PSCD still work well in such circumstances, especially in $\text{mAP}_{75}$, where PSC outperforms KLD by 9.83\,pp on average. Based on the above results, it can be concluded that PSC and PSCD can obtain high performance more easily than CSL and KLD when facing a new dataset.

\begin{table}[t]
\centering
\small
\renewcommand{\arraystretch}{1.2}
\setlength{\tabcolsep}{1.5mm}
\setlength{\abovecaptionskip}{2mm}
\begin{tabular}{l|l|c|c|c}
\hline
\multicolumn{2}{c|}{Method}            & \makebox[14mm][c]{$\text{mAP}_{50}$} & \makebox[14mm][c]{$\text{mAP}_{75}$} & \makebox[14mm][c]{$\text{mAP}_{50:95}$} \\ \hline
\hline
\multirow{5}{*}{\,\rotatebox{90}{FCOS (R-50)}} & \makebox[14mm][l]{Rotated} & 89.74 & 77.00 & 63.84    \\ \cline{2-5}
& +KLD          & 89.76 & 77.46 & 62.63    \\ \cline{2-5}
& +CSL          & 89.84 & 66.47 & 58.92    \\ \cline{2-5}
& +PSC          & {\color{red}\bf90.06} & {\color{blue}78.56} & {\color{blue}67.57}    \\ \cline{2-5}
& +PSCD         & {\color{blue}89.91} & {\color{red}\bf79.20} & {\color{red}\bf67.88}    \\ \hline
\hline
\multirow{5}{*}{\,\rotatebox{90}{\scalebox{0.9}[1.0]{RetinaNet (R-50)}}} & Rotated & 83.50 & {\color{blue}59.60} & 51.56    \\ \cline{2-5}
& +KLD     & {\color{red}\bf85.85} & 58.76 & {\color{blue}53.40}    \\ \cline{2-5}
& +CSL     & 84.87 & 38.75 & 44.17    \\ \cline{2-5}
& +PSC     & {\color{blue}85.65} & {\color{red}\bf61.30} & {\color{red}\bf54.14}    \\ \cline{2-5}
& +PSCD    & 85.53 & 59.57 & 53.20    \\ \hline
\end{tabular}
\caption{Detection accuracy on HRSC dataset.}
\label{tab:hrsc}
\end{table}

\begin{table}[t]
\centering
\small
\renewcommand{\arraystretch}{1.2}
\setlength{\tabcolsep}{1.5mm}
\setlength{\abovecaptionskip}{2mm}
\begin{tabular}{l|l|c|c|c}
\hline
\multicolumn{2}{c|}{Method} & \makebox[14mm][c]{$\text{mAP}_{50}$} & \makebox[14mm][c]{$\text{mAP}_{75}$} & \makebox[14mm][c]{$\text{mAP}_{50:95}$} \\ \hline
\hline
\multirow{5}{*}{\,\rotatebox{90}{FCOS (R-50)}} & \makebox[14mm][l]{Rotated} & {\color{blue} 87.88} & 75.00 & {\color{blue} 64.26}    \\ \cline{2-5}
& +KLD          & 87.72 & 67.41 & 59.77 \\ \cline{2-5}
& +CSL          & 87.23 & 73.82 & 63.12     \\ \cline{2-5}
& +PSC          & {\color{red}\bf 88.87} & {\color{red}\bf 75.72} & {\color{red}\bf 64.85}    \\ \cline{2-5}
& +PSCD         & 87.49 & {\color{blue} 75.48} & 64.18    \\ \hline
\hline
\multirow{5}{*}{\,\rotatebox{90}{\scalebox{0.9}[1.0]{RetinaNet (R-50)}}} & Rotated & 74.68 & {\color{blue} 64.25} & {\color{blue} 55.85}    \\ \cline{2-5}
& +KLD     & {\color{blue} 76.30} & 54.27 & 49.06    \\ \cline{2-5}
& +CSL     & 75.38 & 61.92 & 53.14    \\ \cline{2-5}
& +PSC     & {\color{red}\bf 77.35} & {\color{red}\bf 65.61} & {\color{red}\bf 57.58}    \\ \cline{2-5}
& +PSCD    & 75.77 & 64.24 & 55.70    \\ \hline
\end{tabular}
\caption{Detection accuracy on OCDPCB dataset.}
\label{tab:ocdpcb}
\end{table}

\subsection{Inference time}

With images of $1024 \times 1024$ resolution from DOTA dataset, the average inference time and the model size are evaluated in \cref{tab:time}, which demonstrate that PSC and PSCD are slightly slower than KLD but much faster than CSL.

\begin{table}
\centering
\small
\renewcommand{\arraystretch}{1.2}
\setlength{\tabcolsep}{3mm}
\setlength{\abovecaptionskip}{2mm}
\begin{tabular}{l|c|c}
\hline
\multicolumn{1}{c|}{Method} & Time (ms)     & Parameters   \\ \hline
\multicolumn{1}{c|}{RetinaNet (R-50)} & 42.9 (base) & 36.71M (base) \\ \hline
+KLD              & 43.0 \color{red}\bf{(+0.1)}  & 36.71M \color{red}\bf{(+0.00M)}   \\ \hline
+CSL ($\omega=4$) & 46.3 \color{red}\bf{(+3.4)}  & 37.64M \color{red}\bf{(+0.93M)}   \\ \hline
+CSL ($\omega=1$) & 50.5 \color{red}\bf{(+7.6)}  & 40.44M \color{red}\bf{(+3.73M)}   \\ \hline
+PSC                           & 43.1 \color{red}\bf{(+0.2)}  & 36.79M \color{red}\bf{(+0.08M)}   \\ \hline
+PSCD                          & 43.4 \color{red}\bf{(+0.5)}  & 36.88M \color{red}\bf{(+0.17M)}   \\ \hline
\end{tabular}
\caption{Comparisons on inference time and model size.}
\label{tab:time}
\end{table}

\section{Conclusions}
\label{sec:conclusions}

A novel differentiable angle coder for oriented object detection named phase-shifting coder (PSC) is proposed in this paper, which encodes the orientation angle into periodic phases to solve the boundary discontinuity problem. Upon PSC, an enhanced dual-frequency version (PSCD) mapping the rotational periodicity of different cycles into the phase of different frequencies is proposed to elegantly solve both boundary discontinuity and square-like problems.

Afterward, extensive experiments are carried out to evaluate the performance of PSC and PSCD, through which the following conclusions can be drawn:

1) Among angle-coder-based methods, PSC shows significant improvement in both $\text{mAP}_{50}$ and $\text{mAP}_{75}$ compared with the currently most widely used method CSL. On DOTA dataset, PSCD outperforms CSL by an average of 1.15\,pp in $\text{mAP}_{50}$ and 2.59\,pp in $\text{mAP}_{75}$.

2) Compared with other state-of-the-art approaches, PSC is on a par with Gaussian-distribution-based method KLD in $\text{mAP}_{50}$, but significantly better than KLD in $\text{mAP}_{75}$. In particular, PSCD outperforms KLD by an average of 2.36\,pp in $\text{mAP}_{75}$ on DOTA dataset. When high-IoU bounding boxes are required, PSC and PSCD are expected to give a competitive performance. 

3) The $\text{mAP}_{75}$ of PSCD is considerably higher than that of PSC on DOTA dataset, with the improvement reaching 3.94\,pp on YOLOv5s and 3.60\,pp on YOLOv5m. However, on HRSC and OCDPCB datasets, the dual-frequency strategy plays a negative role, indicating that in scenarios containing rather few square-like objects, PSC could be a better alternative than PSCD.

4) Many existing approaches require different parameters on different datasets. PSC and PSCD, by contrast, do not use dataset-dependent hyper-parameter tuning, so they can be applied to new scenarios more comfortably.

In many scenarios, some objects require ${360}^{\circ}$ oriented detection, while others are ${180}^{\circ}$ or ${90}^{\circ}$ rotationally symmetric. This is an extended form of the square-like problem and a common situation in real applications. In such a case, PSC can still be theoretically applicable, which will be further explored in our future work.

{\small
\bibliographystyle{ieee_fullname}
\bibliography{egbib}
}

\end{document}